\documentclass[journal]{IEEEtran}
\usepackage{cite}
\usepackage{amsmath,amssymb,amsfonts}
\usepackage{graphicx}
\usepackage{textcomp}
 \usepackage{subfigure}
\usepackage[flushleft]{threeparttable} 
\usepackage{booktabs}
\usepackage{algorithm}
\usepackage{algorithmic}

\usepackage[colorlinks = true,
            linkcolor = blue,
            urlcolor  = blue,
            citecolor = blue,
            anchorcolor = blue]{hyperref}
\usepackage{microtype}
\allowdisplaybreaks

\usepackage{url}

\usepackage{comment}

\usepackage{multicol}
\usepackage{multirow}
\usepackage{dsfont}
\usepackage{mathtools}
\usepackage{cuted}
\usepackage[makeroom]{cancel}
\usepackage{stmaryrd}
\usepackage{booktabs, makecell}
\usepackage{pifont}

\usepackage{color}
\definecolor{b2}{RGB}{51,153,255}
\definecolor{p2}{RGB}{121,64,255}

\newcommand{\eg}{{\em e.g.,~}}           
\newcommand{\ie}{{\em i.e.,~}}          
         
\newcommand{\etal}{{\em et al.~}}

\usepackage[T1]{fontenc}

\usepackage{color}
\usepackage{mathtools}

\newcommand{\xmark}{\ding{55}}%

\newcommand{\foot}[1]{{\color{black}#1}}

\renewcommand{\algorithmicrequire}{\textbf{Input:}}  
\renewcommand{\algorithmicensure}{\textbf{Output:}}

\newcommand{\eat}[1]{}

\usepackage{cite}

\begin{document}
%
\title{MISSU: 3D Medical Image Segmentation via Self-distilling TransUNet}
%
%
%

\author{Nan Wang, Shaohui Lin,\IEEEmembership{Member, IEEE}, Xiaoxiao Li, \IEEEmembership{Member, IEEE}, Ke Li, Yunhang Shen, Yue Gao, \IEEEmembership{Senior Member, IEEE} and Lizhuang Ma, \IEEEmembership{Member, IEEE}
\thanks{Corresponding authors: Shaohui Lin, and Lizhuang Ma.}
\thanks{This work was supported in part by the National Natural Science Foundation of China (NO. 72192812, NO. 62102151), Shanghai Sailing Program (21YF1411200), and CAAI-Huawei MindSpore Open Fund (CAAIXSJLJJ-2021-031A).}
\thanks{N. Wang, S. Lin and L, Ma are with School of Computer Science and Engineering, East China Normal University, Shanghai, 200062, China (Email: \{shlin lzma\}@cs.ecnu.edu.cn).}
\thanks{X. Li is with Electrical and Computer Engineering, the University of British Columbia, Vancouver, BC V6T 1Z4  Canada.}
\thanks{K. Li and Y. Shen are with the Youtu Lab, Tencent, Shanghai, 200123, China.}
\thanks{Y. Gao is with School of Software, Tsinghua University, Beijing, 100084, China.}
}

%
%

\markboth{Submission to IEEE TRANSACTIONS ON MEDICAL IMAGING}%
{Shell \MakeLowercase{\textit{et al.}}: Bare Demo of IEEEtran.cls for IEEE Journals}

\maketitle

\begin{abstract}
U-Nets have achieved tremendous success in medical image segmentation. Nevertheless, it may suffer limitations in global (long-range) contextual interactions and edge-detail preservation. In contrast, Transformer has an excellent ability to capture long-range dependencies by leveraging the self-attention mechanism into the encoder. 
Although Transformer was born to model the long-range dependency on the extracted feature maps, it still suffers from extreme computational and spatial complexities in processing high-resolution 3D feature maps. 
This motivates us to design the efficiently Transformer-based UNet model and study the feasibility of Transformer-based network architectures for medical image segmentation tasks.
To this end, we propose to self-distill a Transformer-based UNet for medical image segmentation, which simultaneously learns global semantic information and local spatial-detailed features. Meanwhile, a local multi-scale fusion block is first proposed to refine fine-grained details from the skipped connections in the encoder by the main CNN stem through self-distillation, only computed during training and removed at inference with minimal overhead.
Extensive experiments on BraTS 2019 and CHAOS datasets show that our MISSU achieves the best performance over previous state-of-the-art methods. Code and models are available at \url{https://github.com/wangn123/MISSU.git}
\end{abstract}


\section{Introduction}
\label{introduces}

\begin{figure*}[t]
 \centering
 \includegraphics[width=\linewidth]{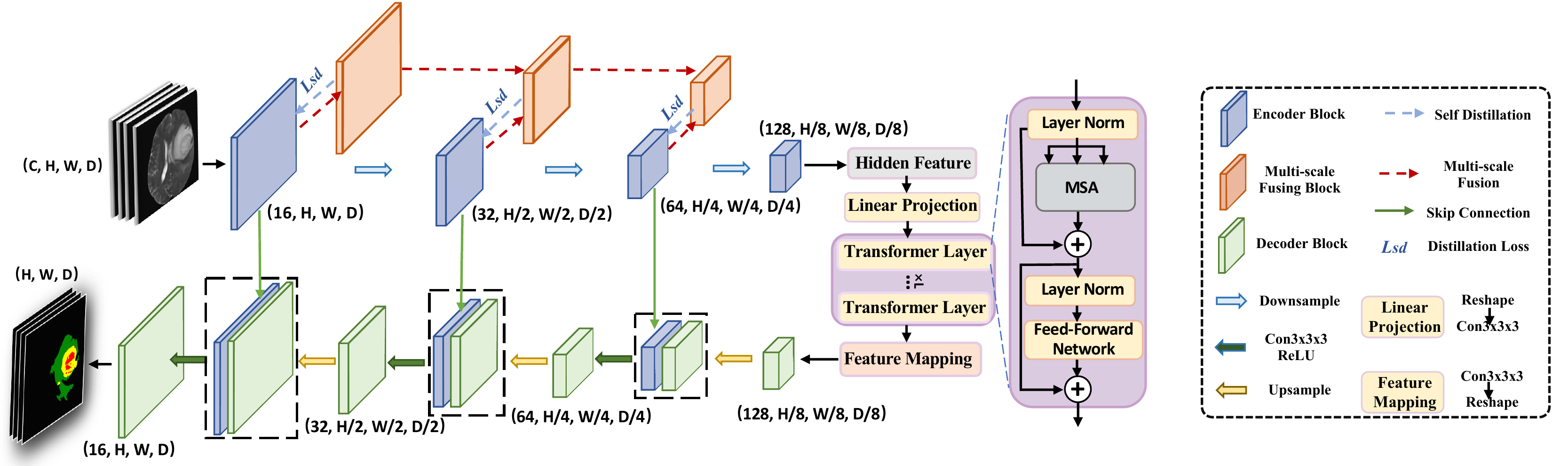}
 \caption{Overview of the proposed MISSU. First, Transformer putting after 3D local features models the global features with long-range dependency. Multi-scale fusion blocks are then constructed like feature pyramids to obtain rich multi-scale detailed features. To reduce the computation cost, 3D local features can be refined by self-distillation to learn the knowledge from multi-scale fusion outputs. Finally, the global features are upsampled and implicitly fused into the refined local features via skip-connection in the decoder to progressively generate the detailed segmentation map. The dotted arrows are only used in training, while the solid ones are for both training and inference. Best viewed in color.
}
 \label{framework}
\end{figure*}

Medical image segmentation plays a vital role in computer-aided diagnosis~\cite{stoitsis2006computer}, which has achieved remarkable success with the usage of convolutional neural networks (CNNs). Among various network architectures, UNet~\cite{ronneberger2015u} has been a mainstream framework for medical image segmentation, as it uses simple yet effective skip-connections to leverage the low-level and high-level semantic features of the encoder into the decoder. Subsequently, substantial efforts have been devoted to further improving the performance for medical image segmentation by using variant UNets, (\eg VNet3D~\cite{milletari2016v}, DAF~\cite{wang2018deep} and nnUNet~\cite{isensee2021nnu}), which have received ever-increasing research attention.


Due to the limited receptive field of UNets, it is difficult to establish an explicit long-distance dependency~\cite{wang2021transbts}.
Inversely, Transformer~\cite{dosovitskiy2020image} makes the self-attention mechanism feasible on a global scale, it can effectively learn long-distance dependency. 
Consequently, researchers improve the feature representation by integrating UNet-based methods with Transformers. 
For example, transformer-based UNets \cite{wang2021transbts,chang2021transclaw}
have been proposed to explicitly capture the long-distance dependency, expanding the limited receptive fields of convolutional kernels in vanilla UNet to learn global semantic information.
Despite the merits they brought, these modules alone may only achieve sub-optimal performance due to the lack of consideration for a compensatory relationship between global and local feature representation.
Especially, it is crucial to learn the global semantic information and detailed local features simultaneously for medical image segmentation. Significantly, the deep branches/layers should extract semantic context information from low-resolution inputs, while the shallow ones concentrate on capturing spatial details (\eg texture and edge) from the high-resolution inputs. Multi-path CNNs~(\cite{mehri2021mprnet,dolz2018hyperdense}) directly fuse the path of global semantic features and the path of local spatial information to improve segmentation performance. However, the computational cost for multi-path inference is significantly high and there is an explicit trade-off between positioning accuracy and multi-path information. 
\textit{This arouses our rethinking: how to design a unified framework for segmentation that implicitly models global semantic contexts and local spatial-detailed information during training while being efficient at inference?}

Inspired by the powerful empirical results of transformer-based models on visual tasks, their promising generalization and robustness characteristics, and their flexibility to model long-range interactions. We propose to self-distill a transformer-based UNet for 3D Medical Image Segmentation (\emph{MISSU}), which \textit{implicitly fuses} the long-distance dependency in global semantic information and the multi-scale local spatial-detailed features and brings no cost during testing. Like UNet, the proposed MISSU is also built upon the encoder-decoder structure with skip connections, whose flowchart is presented in Fig.~\ref{framework}. Specifically,
the encoder first employs 3D CNN to extract features while down-sampling the spatial size. As such, we can capture local 3D context information from the last 3D feature maps.
In this way, rich local 3D context features are effectively embedded in feature representation. 
And then each of them is reshaped into a vector (\ie token) and fed into Transformer layers to model long-range correlations with a global receptive field. 
Additionally, to well extract local information, we design a multi-scale fusion block that receives local 3D features and generates multi-scale fusion outputs. Moreover, the former fusion outputs are progressively aggregated with the current local 3D features to generate the new fusion output. In this way, multi-scale fusion outputs rich detailed features. To relieve the computation burden during the test, we further propose a self-distillation mechanism to transfer the knowledge from multi-scale fusion outputs to local 3D features at the same layers. Self-distillation is formulated by constraining the difference between the local feature and the multi-scale fusion output, which is only used during training. Lastly, the global features are upsampled and implicitly fused into the refined local features via skip-connection in the decoder to progressively generate the detailed segmentation map.


Our main contributions are summarized as follows:
\begin{itemize}

\item We propose a transformer-based UNet variant for medical image segmentation, which can effectively learn global semantic information and local spatial-detailed features. It can implicitly fuse local-to-global features to improve the segmentation performance.

\item Self-distillation is introduced to transfer the knowledge of multi-scale local fusion outputs to local 3D features, which helps local 3D features improve local feature presentation ability. Additionally, self-distillation is computation-free at inference.

\item Extensive experiments demonstrate the superior performance of the proposed MISSU framework for medical image segmentation. On the widely-used BraTS $2019$ dataset, our method achieves $89.98\%$, $85.77\%$ and $80.14\%$ on the segmentation of whole tumor, tumor core and enhanced tumor, respectively, notably outperforming state-of-the-art methods.

\end{itemize}

The remainder of this paper is organized as follows: In Sec. \ref{relate}, related works about medical image segmentation, Transformer and self-distillation learning are introduced. The key components of the proposed MISSU are described in Sec. \ref{method}, such as network encoder (Sec. \ref{nen}), self-distillation (Sec. \ref{Self-distillation}), global feature modeling by Transformer (Sec. \ref{global_transformer}), and network decoder (Sec. \ref{NetworkDecoder}).
Elaborate experiments and analysis are conducted in Sec. \ref{experiments}.
We give the discussion of the proposed MISSU in Sec. \ref{discussion}. Finally, algorithm summary of MISSU is presented in Sec. \ref{conclusion}.

\section{Related work}
\label{relate}

\subsection{Medical image segmentation} 
%
%
As discussed in Sec. \ref{introduces}, local-to-global feature modeling is essential for medical image segmentation. There are various works either enhancing local features or modeling global feature representation~\cite{chen2018drinet}. 

Local feature enhancement is often implemented by image/feature pyramid with multi-scale information fusion~\cite{farabet2012learning}, which has also been applied into medical image segmentation~\cite{feng2020cpfnet}. For example, 
CPFNet~\cite{feng2020cpfnet} designs a scale-aware pyramid fusion (SAPF) module to dynamically fuse multi-scale context information in a top-down manner. MLCFC~\cite{kamnitsas2017efficient} employs a multi-scale 3D CNN to incorporate both local and larger contextual information.
Different from these methods, we propose the multi-scale fusion block to progressively refine the detailed local features in a bottom-up manner, which can be removed to reduce the computation cost by self-distillation at inference.
Similarly, global feature modeling also improves segmentation performance, which is often implemented by global interaction attention~\cite{han2020attention,schlemper2019attention} and self-supervised learning~\cite{zhou2019models}. Firstly, attention mechanism~\cite{wang2018non} can capture the long-distance dependence in the feature map to learn global feature representation. Attention gate model~\cite{schlemper2019attention} can be integrated into U-Net architectures to increase the model sensitivity and segmentation performance. Equally, attention-oriented U-Net model~\cite{han2020attention} replaces convolutional layers with reside-density module via attention mechanism. 
Consequently, they fail to comprehensively capture interactions and similarities between subjects. To tackle this issue,  
we employ Transformer and multi-scale fusion strategies to get local-to-global interaction attention.
Then, self-supervised medical image segmentation aims to learn the global feature representation by designing various pre-text tasks
(\eg solving jigsaw puzzles~\cite{taleb2021multimodal,tao2020revisiting}, rotation prediction~\cite{li2021rotation} and context restoration~\cite{chen2019self, ross2018exploiting}).
However, self-supervised methods require two-stage training (\ie pre-training and fine-tuning), which are computationally intensive for training.
In contrast, our MISSU trains the model from scratch in an end-to-end manner, which is simple yet efficient for training.
%


\subsection{Transformer for medical image}
Transformer as a new attention-driven building block~\cite{dosovitskiy2020image,korkmaz2022unsupervised} has been applied to medical image segmentation, which establishes long-range dependence to capture context information. For example, UNETR~\cite{hatamizadeh2021unetr} utilizes the pure transformer as the encoder to learn sequence representations of the input volume, which captures the global multi-scale information. TransUNet~\cite{chen2021transunet} employs the transformer to process each 3D medical image in a slice-by-slice manner, which cannot well learn continuous information between slices. To better use temporal information, TransBTS~\cite{wang2021transbts} uses 3D CNNs to extract the context information and employs a Transformer to model global feature representation with long-range dependence. 
Differently, our MISSU simultaneously learns global semantic information and refine local features to generate spatial-detailed features, further improve the segmentation performance.

Equally, multi-path fusion learning between local and global features~\cite{chandrakar2020brain,wei2020multi} has also been applied to medical image segmentation. For example, Vessel-Net~\cite{wu2019vessel} introduces multi-paths to preserve the rich and multi-scale deep features during model optimization. D-MEM~\cite{zhao2019automated} proposes a deformable multi-path ensemble for both local and global features for automated cervical nuclei segmentation.
However, multi-path fusion is computation-intensive, as all local and global paths need to be computed during inference. Differently, we employ implicit fusion between local and global features by self-distillation so that the computation can be removed at inference.


\subsection{Self-distillation learning} 
Self-distillation explores the potential of knowledge distillation from a new perspective~\cite{qin2021efficient}. Similarly,
self-distillation strategies can be summarized as extracting the attention map of the current layers and then transferring the knowledge to the previous layer~\cite{zhang2021cross}. It aims to improve the performance of a compact model by using its knowledge without a teacher network.
Self-distillation methods~\cite{ahn2019variational,yang2019training} are mostly applied to language modeling, image recognition and object detection. 
For example, Hou \etal~\cite{hou2019learning} presented a self-attention distillation approach for lane detection by allowing a network to utilize the attention maps of its own layers as distillation targets for its following layers.
To our best knowledge, it is unexploited for medical image segmentation, except KD-ResUNet++~\cite{kang2020kd}. However, KD-ResUNet++ uses past predictions about data from the model to soften the targets at the previous epoch. It has a large number of parameters and is overly dependent on the previous period model. 
Inspired by previous work, we employ the self-distillation
framework in medical image segmentation, and improve the method so that it can learn the student network by transferring the knowledge online from its auxiliary information. 



%

\section{Method}
\label{method}

\subsection{Overall pipeline of MISSU}
As illustrated in Fig.~\ref{framework}, the proposed network architecture of MISSU is built on the encoder-decoder framework.
In the encoder, an input MRI scan $\mathcal{X}\in\mathbf{R}^{C\times H\times W\times D}$ with $C$ channels (modalities), $H\times W$ spatial resolution and $D$ depth dimension (slices) first goes through a 3D CNN-based encoder to generate local feature maps, capturing spatial and depth information. Then, Transformer layers are used to model global feature representation with long-distance dependency. To compensate for the feature with detailed local information (\eg shape and border of organs), we introduce the multi-scale fusion block that receives local features and generates multi-scale fusion outputs. Furthermore, self-distillation is proposed to transfer the knowledge from multi-scale fusion outputs to local features at the same layers during training, which can be removed at inference to reduce the computation cost. Finally, the decoder consists of multiple upsampling steps, decodes hidden features containing global and local information, and progressively produces the full resolution segmentation map. The detailed network architecture of MISSU is provided in our released code\footnote{https://github.com/wangn123/MISSU.git}.

\subsection{Network encoder}
\label{nen}
Since the computing complexity of Transformer is polynomial with respect to the amount of tokens (\ie sequence length), simply flattening input image to sequence as Transformer input is unfeasible. For example,
ViT~\cite{dosovitskiy2020image} applies Transformer on computer vision, tokens are often constructed by splitting an image into patches and fed into self-attention blocks to model long-range feature dependency. Specifically,
ViT divides an image into fixed-size $16 \times 16$ patches, then reshapes each patch into a token, decreasing the sequence length to $16 \times 16$.
For 3D volumetric input data with spatial and depth dimensions, the naive tokenization method will not model the detailed local information~\cite{yuan2021tokens} across spatial and depth dimensions for volumetric segmentation. 

To address the above problem, we first apply $3 \times 3 \times 3$ convolution blocks with downsampling to gradually encode input image into a low-resolution/high-level feature representation $\mathcal{A}^s\in\mathbf{R}^{N\times \frac{H}{2^{s-1}} \times \frac{W}{2^{s-1}} \times \frac{D}{2^{s-1}}}$ with channels $N=16*2^{s-1}, s=1,\cdots, 4$, where $s$ is the stage number. As such, we can extract the local detailed features in $\mathcal{A}^s$. Subsequently, the input $\mathcal{A}^4$ is fed into Transformer to learn the long-range correlation across spatial and depth dimensions with the global receptive field.

\textbf{Transformer block.} 
We first construct tokens at the local feature $\mathcal{A}^4\in\mathbf{R}^{128\times\frac{H}{8}\times\frac{W}{8}\times\frac{D}{8}}$. We reshape $\mathcal{A}^4$ into a sequence of volumetric vector/token $\mathbf{a}_p\in\mathbf{R}^{128}$ according to the total length $M$ of spatial and depth dimension $D$, where $p=1,2,\cdots,M$ (\emph{i.e.,} $\frac{H}{8}\times\frac{W}{8}\times\frac{D}{8}$). To ensure a comprehensive presentation of each volume in $\mathbf{a}_p$, linear projection with weight $\mathbf{E}$ is used to increase the channel dimension $N$ to $d$ in the embedding space. Following ViT~\cite{dosovitskiy2020image}, we also introduce specific learnable position embedding $\mathbf{E}_{pos}$, which is added into embedding tokens for retaining positional information. Different with ViT, we remove the class token for segmentation task. Therefore, we can formulate the above computation as:
\begin{equation}
\label{trans}
\mathbf{Z}_{0}=\left[\mathbf{a}_{p}^{1} \mathbf{E} ; \mathbf{a}_{p}^{2} \mathbf{E} ; \cdots ; \mathbf{a}_{p}^{M} \mathbf{E}\right]+\mathbf{E}_{pos},
\end{equation}
where $E\in\mathbf{R}^{512\times128}$ and ${E}_{pos}\in\mathbf{R}^{M\times512}$ are the patch embedding projection and the position embedding, respectively. $\mathbf{Z}_{0}\in\mathbf{R}^{M\times512}$ denotes the output feature embedding.

We then feed the feature embedding outputs $Z_{0}$ into the Transformer encoder, which consists of $L$ Transformer layers. A Transformer layer comprises a multi-head self-attention (MSA) block, and a feed-forward network (FFN). The output $\mathbf{Z}_{l}$ in the $l$-th Transformer layer is defined as follows:
\begin{equation}
\label{eq2}
\begin{split}
& \mathbf{Z}_{l}^{\prime}=MSA\left(LN\left(\mathbf{Z}_{l-1}\right)\right)+\mathbf{Z}_{l-1}, \quad l=1,2,\cdots, L \\
& \mathbf{Z}_{l}=FFN\left(LN\left(\mathbf{Z}_{l}^{\prime}\right)\right)+\mathbf{Z}_{l}^{\prime}, \quad l=1,2,\cdots, L,
\end{split}
\end{equation}
where $LN(\cdot)$ is the layer normalization.

Finally, the feature mapping consists of one linear project layer and reshape operation, which is applied to the Transformer output $\mathbf{Z}_{L}$ to generate the global features $\mathcal{Z}^4$ with the same size as local feature $\mathcal{A}^4$. 

\subsection{Detailed Feature Modeling by Self-distillation}
\label{Self-distillation}

To successfully segment different lesion areas with small and significant shapes, the local features $\mathcal{A}^s$ should be enhanced to extract multi-scale local detailed information. Therefore, we proposed multi-scale fusion block (MSF) and self-distillation to obtain multi-scale local detailed feature preservation for computation-free inference.

\begin{figure}[t]
 \centering
 \includegraphics[width=\linewidth]{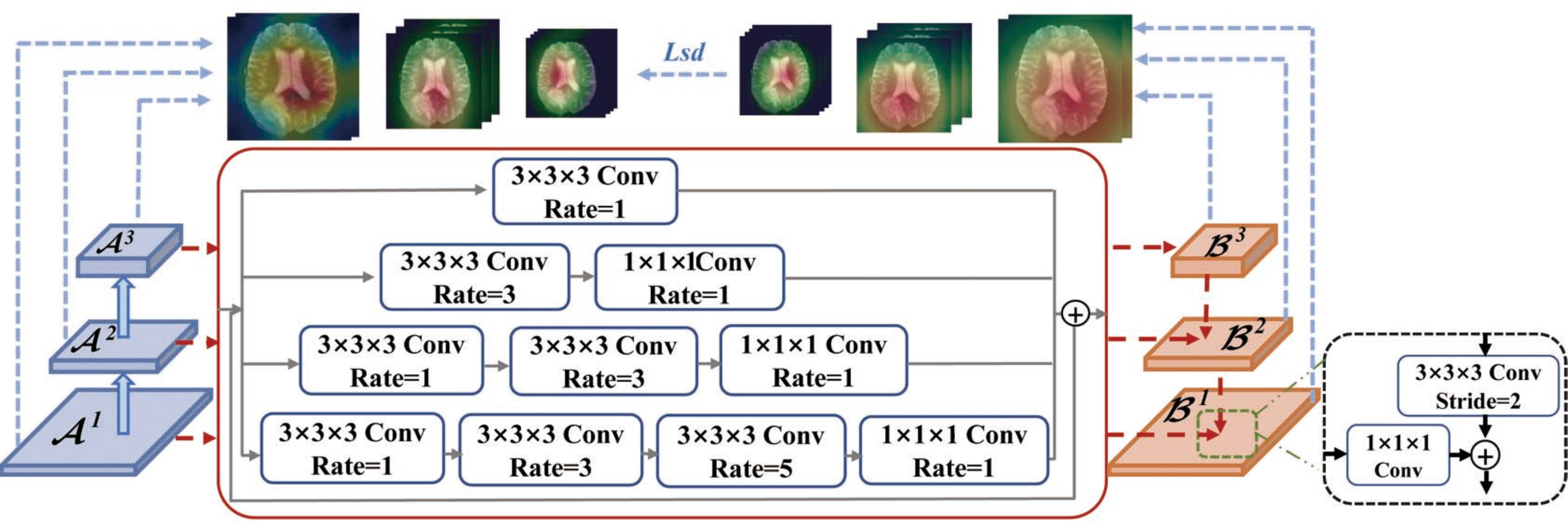}
 \caption{The flowchart of the multi-scale fusion (MSF) and self-distillation. MSF contains four cascade branches, each branch is implemented by a serial of atrous convolutions to achieve different receptive field of 3, 7, 9 and 19.
 In each atrous branch, we produce the same size as the local features. As such, we add the feature maps from different branches for feature fusion.}
 \label{msf}
\end{figure}

\textbf{Multi-scale fusion block.}
%
Inspired by deeplabv3~\cite{chen2017rethinking}, we propose a novel multi-scale fusion block (MSF), which is leveraged as a feature pyramid in a bottom-up manner. 
As shown in Fig.~\ref{msf}, the proposed MSF is different from the Atrous Spatial Pyramid Pooling (ASPP) module~\cite{chen2017rethinking} based on two aspects: On the one hand, we add our MSF on the shallow and intermediate features rather than features from the top layer, which makes the local features learn multi-scale detailed information for segmentation; On the other hand, the pooling layer in ASPP is removed to reduce the information loss of detailed features. We also introduce a feature pyramid to construct the multi-level fused features at the bottom-up pathway rather than the top-down one in FPN \cite{lin2017feature}.

Input local features $\mathcal{A}^s, s=1,2,3$ are first fed into MSF to generate multi-scale refined local feature maps $\bar{\mathcal{A}}^s, s=1,2,3$, which utilizes four parallel branches $f_p^{s}$ with parameters $\theta_p^{s} (p=1,\cdots,4)$.  
Each branch is implemented by a serial of atrous convolutions to achieve the different receptive fields of 3, 7, 9 and 19. In particular, the first branch uses one $3\times 3\times3$ atrous convolution with an atrous rate of 1 to generate feature maps with the receptive field of 3. The second branch uses one $3\times 3\times3$ atrous convolution with an atrous rate of 3 following by one $1\times 1\times1$ regular convolution, which can generate feature maps with the receptive field of 7. Compared to the second branch, the third branch increases one more $3\times 3\times3$ atrous convolutions with an atrous rate of 1, which obtains the receptive field of 9. The fourth branch also has one more $3\times 3\times3$ atrous convolutions to produce the receptive field of 19, compared to the third one. 
The total four branches' outputs $f_p^{s}(\mathcal{A}^s;\theta_p^{s})$ and input local features $\mathcal{A}^s$ are then added to generate $\bar{\mathcal{A}}^s$. Then, feature pyramids are used to generate multi-level MSF outputs in a bottom-up manner, which can be formulated as:
\begin{equation}
\label{msfblockloss}
\mathcal{B}^1 = \bar{\mathcal{A}}^1; \mathcal{B}^s = \mathcal{W}_{A}^{s}\otimes\bar{\mathcal{A}}^s + \mathcal{W}_{B}^{s}\otimes\mathcal{B}^{s-1}, s=2,3,
\end{equation}
where $\mathcal{B}^s$ is the refined output of local feature at the $s$-th stage. $\mathcal{W}_{A}^{s}$ and $\mathcal{W}_{B}^{s}$ are 3D $1\times1\times1$ and $3\times3\times3$ convolution kernels with strides of 1 and 2, respectively.  

\subsection{Global Feature Modeling by Transformer}
\label{global_transformer}

\textbf{Self-distillation.}
To further reduce the computation of multi-scale fusion blocks for inference, we propose a self-distillation mechanism to transfer the knowledge from the MSF outputs $\mathcal{B}^s$ to its corresponding local features $\mathcal{A}^s$ during training, and the computation for MSF outputs would be skipped. Inspired by~\cite{komodakis2017paying}, we construct attention maps over spatial and depth dimension performing on $\mathcal{B}^s$ and $\mathcal{A}^s$ by attention mapping $\phi$, which can be defined by:
\begin{equation}
\label{eq4}
\phi:\mathbf{R}^{N\times\frac{H}{2^{s-1}} \times \frac{W}{2^{s-1}} \times \frac{D}{2^{s-1}}}\rightarrow \mathbf{R}^{\frac{H}{2^{s-1}} \times \frac{W}{2^{s-1}} \times \frac{D}{2^{s-1}}}.
\end{equation}

In this paper, we use sum of absolute values, \emph{i.e.,} $\phi(\mathcal{A}^s) = \sum_{i=1}^{N}|\mathcal{A}_i^s|$. Hence, the distillation loss can be formulated as:
\begin{equation}
\label{eq5}
\mathcal{L}_{sd} = \sum_{s=1}^{3}\|\frac{q_{\mathcal{A}}^{s}}{\|q_{\mathcal{A}}^{s}\|_2}-\frac{q_{\mathcal{B}}^{s}}{\|q_{\mathcal{B}}^{s}\|_2}\|_2,
\end{equation}
where $q_{\mathcal{A}}^{s} = \textit{vec}(\phi(\mathcal{A}^s))$ and $q_{\mathcal{B}}^{s} = \textit{vec}(\phi(\mathcal{B}^s))$ are local feature $\mathcal{A}^s$ and multi-scale fusion output $\mathcal{B}^s$ attention maps at the $s$-th stage, respectively. Note that we add self-distillation on the three stages, which is due to the lack of local detailed information at the output of the fourth stage.


\begin{algorithm}[t]
\small
\label{alg1}
\caption{The training process of MISSU.}
\renewcommand{\algorithmicrequire}{\textbf{Input:}} 
\renewcommand{\algorithmicensure}{\textbf{Output:}}
\begin{algorithmic}[1]
\REQUIRE 
Training data $\mathcal{D} = \{\mathcal{X},\mathbf{y}\}$, hyper-parameter $\lambda$, maximum iterations $T$.
\ENSURE 
The parameters $\theta_{e},\theta_p^s,\theta_{d}$.\\
\STATE
Initialize $t = 0$ and parameters $\theta_{e},\theta_p^s,\theta_{d}$.
\REPEAT 
\STATE 
 \textbf{Forward Pass:}\\ 
 Choose a minnibatch from $\mathcal{D}$, conduct forward propagation and loss computation with $\theta_{e}$, $\theta_p^s$ and $\theta_{d}$ via Eq.~{\ref{eq6}}.
 \STATE 
 \textbf{Backward Pass:}\\
 Compute the gradient of $\nabla\theta_{e},\nabla\theta_p^s$ and $\nabla\theta_{d}$ by the deviation of Eq.~{\ref{eq6}} with respect to $\theta_{e}$, $\theta_p^s$ and $\theta_{d}$, Respectively.
  \STATE 
 \textbf{Update:}\\
 Update $\theta_{e}$, $\theta_p^s$ and $\theta_{d}$ by Adam with the poly learning rate strategy.
  \STATE 
 $t:=t+1$.
\UNTIL{Convergence or $t$ reach maximum iterations {$T$}.}
\end{algorithmic}

\end{algorithm}

\subsection{Network Decoder}
\label{NetworkDecoder}

In the decoder, skip-connection and up-sampling operation implement the implicit fusion between global features extracted by the Transformer output $\mathcal{Z}^4$ and the local detailed features $\mathcal{A}^s$.
In particular, we first use $2\times$ upsampling operator on $\mathcal{Z}^4$ to generate the upsampling outputs, which are concatenated by $\mathcal{A}^3$ and then followed by one $3\times3\times3$ convolutional layer and one ReLU layer to produce decoder outputs at the 3rd stage. We progressively process the decoder outputs by upsampling, concatenation, convolution and ReLU operator to generate the pixel-level segmentation map.

Overall, we integrate self-distillation loss Eq.~\ref{eq5} with the segmentation loss to construct the overall loss function as:
\begin{equation}
\label{eq6}
\mathcal{L}(\mathcal{X},\mathbf{y};\theta_{e},\theta_p^s,\theta_{d}) = \mathcal{L}_{seg}(\mathcal{X},\mathbf{y};\theta_{e},\theta_p^s,\theta_{d}) + \lambda\mathcal{L}_{sd},
\end{equation}
where $\mathcal{L}_{seg}(\mathcal{X},y;\theta_{e},\theta_p^s,\theta_{d})$ is the traditional segmentation loss based on cross-entropy with encoder parameters $\theta_{e}$, MSF parameters $\theta_p^s$ and decoder parameters $\theta_{d}$. $\mathcal{X}$ and $y$ are input MRI scan images and their corresponding labels, respectively. $\lambda$ is a hyper-parameter to balance segmentation loss and self-distillation loss. Eq.~\ref{eq6} can be minimized by Adam optimizer~\cite{kingmaadam} in Alg. 1.

\section{Experiment}
\label{experiments}
\subsection{Experimental Setups}

\textbf{Datasets.} We experiment on two widely-used medical image segmentation benchmarks: \textit{BraTS 2019 dataset}~\cite{bakas2018identifying} for brain tumor segmentation and \textit{CHAOS dataset}~\cite{kavur2019chaos} for liver segmentation.

\begin{figure*}[t]
\centering
\subfigure[Brain tumor segmentation on BraTS 2019]{%
\includegraphics[width=0.95\columnwidth]{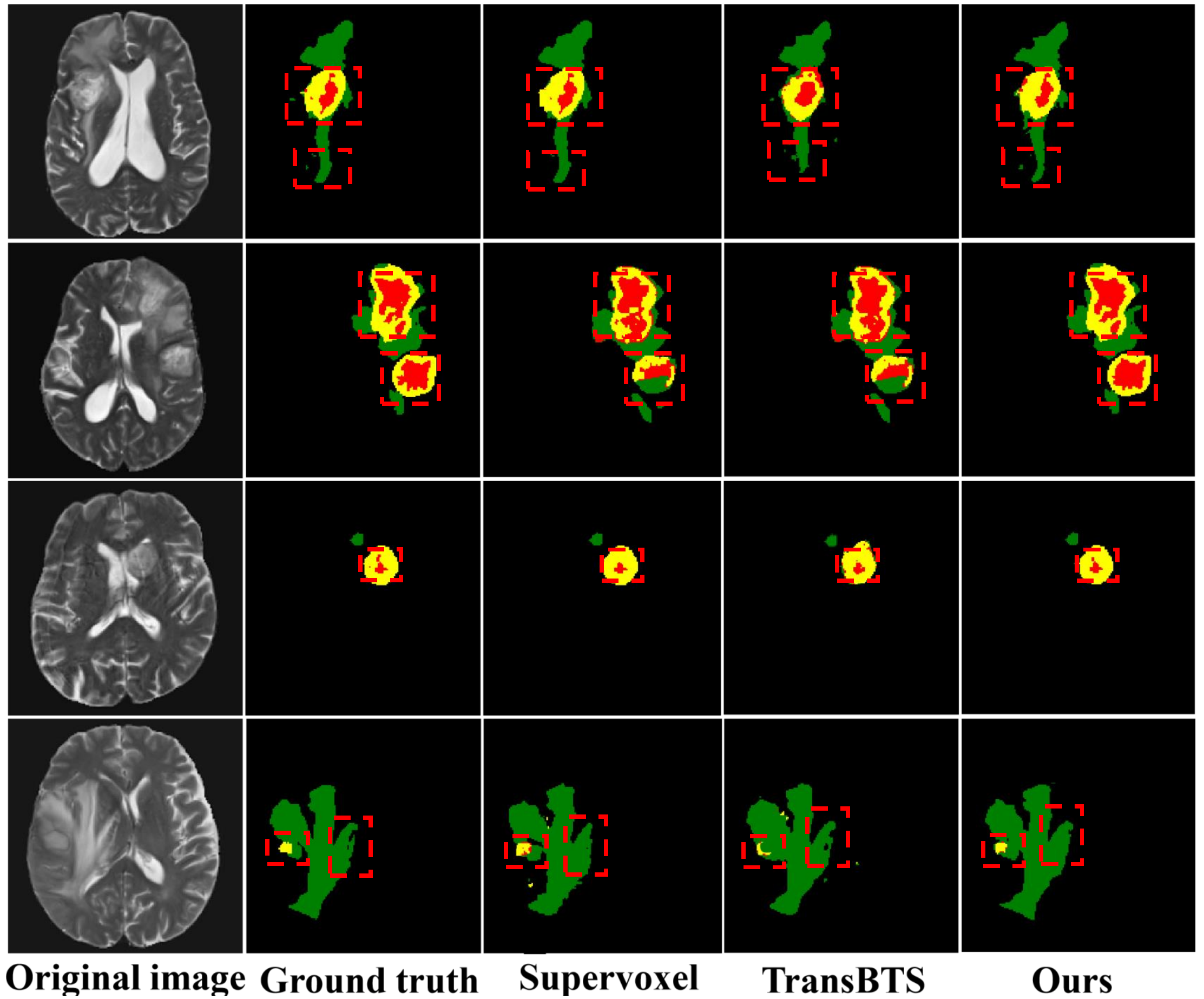}
\label{fig:subfigure1}}
\quad
\subfigure[Liver segmentation on CHAOS dataset]{%
\includegraphics[width=0.95\columnwidth]{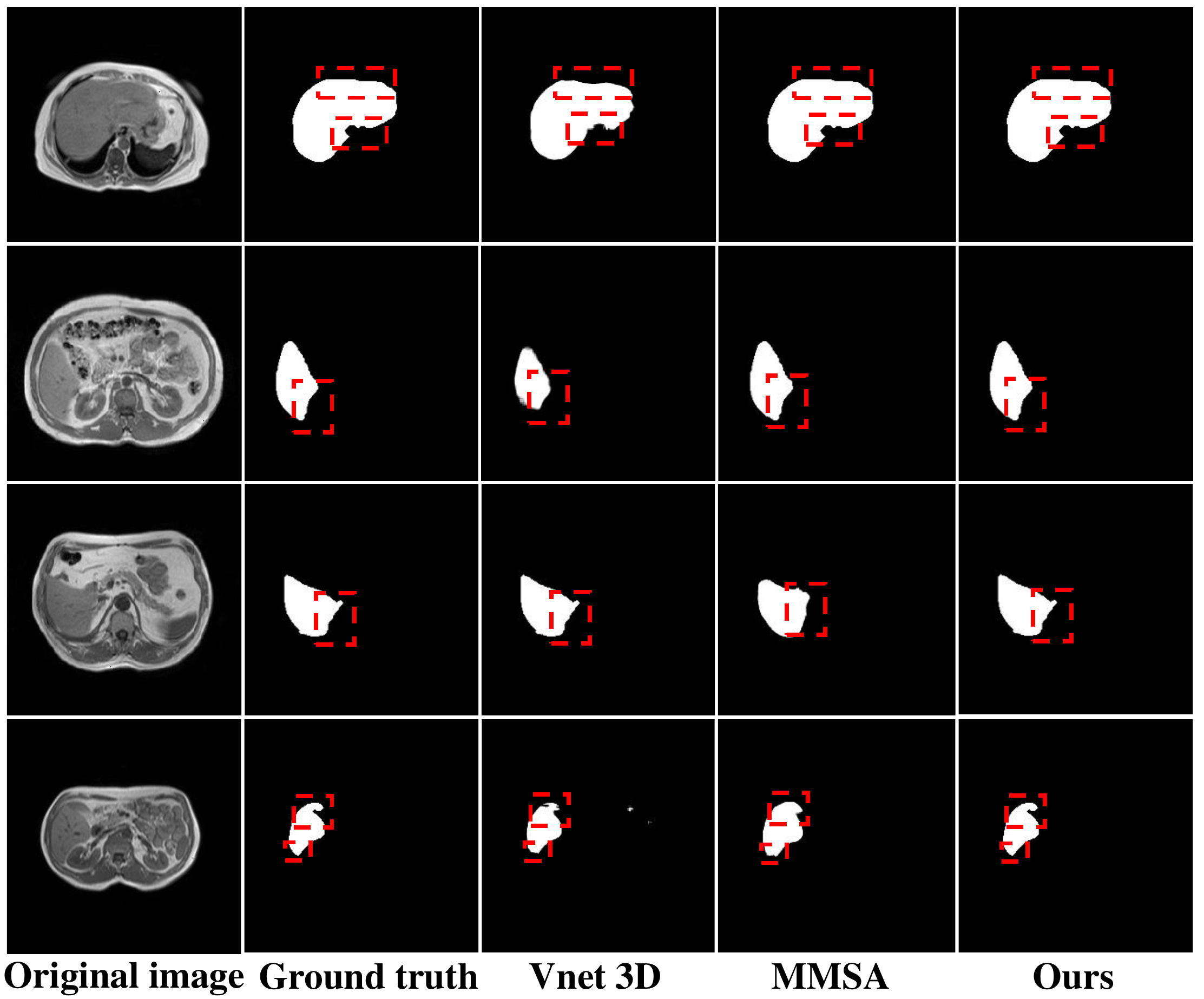}
\label{fig:subfigure2}}
\caption{Visual comparison for brain tumor segmentation on BraTS 2019 and liver segmentation on CHAOS. (a) Brain tumor segmentation results were yellow, green and red areas present enhancing tumor (ET), peritumoral edema (ED) and the necrotic and non-enhancing tumor core (NCR/NET), respectively. (b) Liver segmentation results, where white and black areas denote liver and background/other organs, respectively.}
\label{fig:figure}
\end{figure*}

\textit{BraTS 2019} contains $259$ high-grade glioblastomas (HGG) patients and $76$ low-grade glioblastomas (LGG) patients for training, and $125$ cases for validation. 
Each patient has four image modalities, including T1-weighted (T1), post-contrast T1-weighted (T1ce), T2-weighted (T2) and Fluid Attenuated Inversion Recovery (FLAIR). 
Each modality has been aligned into the same area, which has a volume of $240\times240\times155$. However, intensity normalisation is required for input data to ensure that the grey values of each image have the same distribution. Within each modality, z-score normalisation is used for the foreground with non-zero voxel values in the medical images and corresponding labels.
There are four different ROIs/classes of brain tumors: whole tumor (WT), tumor core (TC), enhanced tumor (ET), and background without tumor. As shown in Fig.~\ref{fig:subfigure1}, WT contains peritumoral edema (green part), enhancing tumor (yellow part), and the necrotic and non-enhancing tumor core (red part); TC means the yellow and red region; ET stands for the red region; background without tumor shows in black background. 

\textit{CHAOS} dataset contains 40 two-modality (\ie T1 and T2) image sequences, where 20 image sequences are used for training and the remaining for testing without annotations. Each sequence has 30-50 slices with a resolution of $256\times256$. Since testing labels are not provided, we follow MMLAO~\cite{frangi2018medical} by splitting the training set into subsets of 15 and 5 subjects for training and testing. As shown in Fig.~\ref{fig:subfigure2}, there are only two ROIs for liver segmentation: liver (white part) and background/other organs (black part).

\textbf{Evaluation metric.} 
For the quantitative analysis of experimental results, we consider multiple performance measurements, including Dice-score, Hausdorff distance and Accuracy (ACC), which are widely used as a criterion for medical image segmentation.
 Dice-score measures the overlap between two samples on the target area, which can be calculated as:
 \begin{equation}
\operatorname{Dice}(\mathrm{P}, \mathrm{T})=\frac{2 \cdot|P \cap T|}{|P|+|T|},
\end{equation}
where $P$ and $T$ indicate the predicted region and the ground truth region, respectively. 
The larger Dice-score, the better the segmentation performance. 
We employ Dice-score to evaluate the segmentation performance for WT, TC and ET on BraTS $2019$ and the liver region on CHAOS. Dice ranges from zero to one that the higher score is better.

Similarly, the ACC is calculated as the ratio of correctly identified samples to total samples. 
The calculation formula is as follows:

\begin{equation}
\text { ACC }=\frac{T P+T N}{T P+T N+F P+F N}~,
\end{equation}
where true positives (TP) and true negatives (TN) represent the number of correctly classified as positives and negatives, respectively.
False positives (FP) and false negatives (FN) are the number of false positives and false negatives, respectively.

To calculate the distance between segmentation boundaries, the Hausdorff distance (HD) is utilized. The highest value of the shortest least square distance ($d(p, t)$) between all points on the predicted label surface ($PL$)  and the points on the ground truth label ($GT$) is calculated. 
\begin{equation}
H D(PL, GT)=\left\{\sup _{p \in \partial p} \inf _{t \in \partial t} d(p, t), \sup _{t \in \partial t} \inf _{p \in \partial p} d(t, p)\right\}.
\end{equation}

\textbf{Implementation details.} We implement our approach in PyTorch $1.7.0$ with 8 NVIDIA 2080Ti GPUs. The models are trained using Adam optimizer by setting $\beta_1=0.9, \beta_2=0.999$, and $\epsilon=10^{-5}$. The learning rate is initialized by $0.0004$, decaying by each iteration with power 0.9 using the poly learning rate strategy. The batch size and total epochs are set to 4 and 1000, respectively. The balance parameter $\lambda$ is empirically set to 0.3. 

In the training phase, the image sequences in the BraTS are randomly cropped into patches from $240\times 240\times 155$ to $128\times128\times128$, 
while the padding procedure may be required to maintain a consistent number of slices within each training case.
And the image sequences augmented with random horizontal flip across the axial, coronal and sagittal planes by a probability of $0.5$. The input channel (modality) $C$ is set to 4. Furthermore, random intensity shift between [-0.1, 0.1] and scale between [0.9, 1.1]. For CHAOS dataset, we randomly cropped its sequences into patches from $256\times256\times30$ to $128\times128\times30$, and $C$ is set to $2$. A number of data augmentation techniques are used during training: rotations, scaling, Gaussian noise, which remain consistent with nnUNet~\cite{isensee2021nnu}.

\subsection{Quantitative and Qualitative Results}

\subsubsection{Comparison methods}
We compare with UNet-based methods, 
knowledge distilled method, 
self-supervised method 
and Transformer-based methods. 
We adopt the open-source codes of all comparison methods and use grid search to decide their best hyper-parameters.
We list details of each comparison method below.
\begin{itemize}
\item \textit{UNet-based methods} (\eg UNet~\cite{ronneberger2015u}, 3D UNet~\cite{cciccek20163d}, nnUNet~\cite{isensee2021nnu}, DAF~\cite{wang2018deep} and VNet3D~\cite{milletari2016v}): the architectures consist of a contracting path, collecting context and a symmetric expanding path, exact localization. 

\item \textit{KD-Net}~\cite{hu2020knowledge} utilizes knowledge distillation to transfer knowledge from a trained multi-modal network (teacher) to a mono-modal one (student).

\item \textit{Supervoxel}~\cite{kayal2020region} generates images to inpaint by employing self-supervised learning-based masking instead of random masking, and also focuses on the region-of-interest for brain tumor segmentation. 

\item \textit{Transformer-based methods} (UNETR~\cite{hatamizadeh2021unetr}, MeTrans~\cite{jun2021medical}, TransUNet~\cite{chen2021transunet}, TransBTS~\cite{wang2021transbts}), MSSA \cite{sinha2020multi},  TransClaw~\cite{chang2021transclaw}).
UNETR, MeTrans, TransUNet and TransBTS are discribed in Sec.~\ref{relate}.
MSSA uses a guided self-attention mechanism to capture contextual dependencies. 
%
TransClaw integrates Transformer layers in the encoding part to exploit multi-scale information. The more compared architectures are presented in Sec.~\ref{discussion_msf}.
\end{itemize}

\begin{table}[t]
\centering
\caption{Quantitative comparison on BraTS 2019 dataset. \textbf{Bold} and \underline{underline} number are the best and ranking-second best performance in all tables, respectively.}
\label{compare}
\begin{tabular}{cccc}
\hline Method  &WT(\%) &TC(\%) &ET(\%)\\
\hline
3D~UNet('16)~\cite{cciccek20163d}   &87.38 &72.48&70.86\\
nnUNet('21)~\cite{isensee2021nnu}    &\underline{89.67} &\underline{84.01}&\underline{78.55}\\
KD-Net('20)~\cite{hu2020knowledge}  & 78.52& 82.70 & 72.89\\
Supervoxel('20)~\cite{kayal2020region} &85.11 &79.34 &72.15\\
UNETR('21)~\cite{hatamizadeh2021unetr} & 79.00&75.82&60.62\\
MeTrans('18)~\cite{jun2021medical} & 87.33&74.39&63.19\\
TransUNet('21)~\cite{chen2021transunet} &89.48&78.91&78.17\\
TransBTS('21)~\cite{wang2021transbts} & 88.89& 81.41& 78.36\\ \hline
 MISSU(Ours) &\textbf{89.98}& \textbf{85.77} &\textbf{80.14}\\
\hline
\end{tabular}
\end{table}

\textbf{Quantitative comparison.}
To begin with, we evaluate the segmentation performance on BraTS 2019, which is summarized in Tab.~\ref{compare}.
Particularly, our method improves by $3.29\%$ on average, compared to the vanilla UNets (\ie 3D UNet, nnUNet). 
This suggests that UNet-based approaches are effective. However, they have limitations in long-distance context fusion, which may lose detailed information, such as edge and texture. 
In addition, KD-Net~\cite{hu2020knowledge} and Supervoxel~\cite{kayal2020region}, employ knowledge distillation and self-supervised learning for brain tumor segmentation, respectively. MISSU outperforms KD-Net by 3.07\% on TC.
Compared with Supervoxel, we obtain the 
result of $85.77\%$, which is higher than that (\ie $80.94\%$) of the TC-Dice.
To this end, MISSU also achieves better performance than the best Transformer competitor TransBTS~\foot{\footnote{TransBTS: \url{https://github.com/Wenxuan-1119/TransBTS}}}.
Specifically, compared to TransBTS, MISSU achieves an average improvement of 1.09\%, 4.36\% and 1.78\%, in terms of WT, TC and ET, respectively.
These results suggest that the effectiveness of our proposed MISSU self-distilling Transformer design can achieve better feature learning performance than the SOTA TransBTS.

We further evaluate the performance of liver segmentation on the CHAOS dataset, which is summarized in Tab.~\ref{liverseg}.
Our MISSU always outperforms the backbones and Transformer-based methods, especially with a large gap in performance on baselines. 
Specifically, our method achieves $14.30\%$ and $6.63\%$ Dice-score gains over DAF \cite{wang2018deep} and VNet3D~\cite{milletari2016v}, whose backbones of the encoder are built by 3D CNNs. 
Compared to Transformer-based TransClaw~\cite{chang2021transclaw}, and our method also achieves better performance for live segmentation with $2.47\%$ Dice-score gains. 
These results confirm our claim that MISSU obtains satisfying prediction performance with local-to-global modeling by self-distillation.

\begin{table}[t]
\centering
\caption{Quantitative comparison on CHAOS dataset.}
\label{liverseg}
\scalebox{1}{
\begin{tabular}{cc}
\hline Method & Dice(\%) \\
\hline
UNet('15)~\cite{ronneberger2015u} & 81.14 \\
VNet3D('16)~\cite{milletari2016v} &90.15\\
DAF('18)~\cite{wang2018deep}  &82.48 \\
MSSA('20)~\cite{sinha2020multi}  &86.75\\
nnUNet('21)~\cite{isensee2021nnu}&93.45\\
TransClaw('21)~\cite{chang2021transclaw} & \underline{94.31}\\ \hline
MISSU(Ours)  &\textbf{96.78}  \\
 \hline
\end{tabular}}
\end{table}

\textbf{Qualitative comparison.}
We collect author-verified codes for all comparison methods and apply the parameter settings recommendations in the associated literature to ensure that all comparison methods operate effectively on each dataset.

Fig.~\ref{fig:subfigure1} shows visual comparisons for brain tumor segmentation with different methods. 
MISSU achieves the best visual segmentation results compared to other baselines (\eg Supervoxel and TransBTS). In particular, the necrotic predicted non-enhancing tumor core (red area) is correctly classified at the exact position and range, as shown in the second row. Moreover, the proposed MISSU can perform better on the local details in the fourth row to enhance tumor edge segmentation and improve each voxel classification. 
Fig.~\ref{fig:subfigure2} presents some liver visual segmentation results, where the liver is marked in white. Compared to VNet3D~\cite{milletari2016v} and MSSA~\cite{sinha2020multi}, our method achieves the best accurate segmentation for contour details. For example, the proposed MISSU can accurately segment the liver boundary in the last row, achieving almost the same results as the ground truth. 
The main reason could be that MISSU can help extract robust features by local-to-global fusing.
Consequently, our experiments show promising
results in tackle the issue of long-range dependencies by leveraging the self-attention mechanism and weight sharing.

\textbf{Model complexity comparison.}
For fairness, both comparative and our methods adopt the same input setting ($4\times128\times128\times128$) to evaluate the GFLOPs and parameters. The results are summarized in Tab.~\ref{tab:flop}. 
Obviously, MISSU significantly outperforms the 3D UNet with 16.32M parameters and 1,670.00 GFLOPs in the computational complexity.
Additionally, compared with TransUNet, although TransUNet achieves the lowest GFLOPs, MISSU reduces $85.57M \downarrow$ in parameters while significantly improving model performance of WT, TC and ET Dice-scores shown in Tab.~\ref{compare}.
The SOTA method TransBTS is a moderate-size model with 208.00 GFLOPs and 15.14M parameters, MISSU only requires 132.11 GFLOPs and 10.50M parameters while achieving the higher Dice scores of $89.98\%$, $85.77\%$, $80.14\%$. Therefore, our method achieves the best trade-off between the segmentation performance and model complexity. To explain, our MISSU can effectively learn local-to-global features by implicitly fusing the global semantic information and multi-scale local spatial-detailed features via self-distillation.  

\begin{table}[t]
\centering
\caption{Complexity comparison between baselines and Transformer-based methods on BraTS 2019. }
\label{tab:flop}
\begin{tabular}{ccc}
\hline
Metric & GFLOPs  & \#Parameters(M) \\ \hline
3D UNet & 1,670.00 & 16.32 \\
nnUNet & 412.65 & 19.07\\
TransUNet   &\textbf{48.34}  &  96.07 \\
TransBTS & 208.00  & \underline{15.14}   \\
UNETR &193.5&102.5\\
\hline
\begin{tabular}[c]{@{}c@{}}Trans.+MSF\\ (MS-output)\end{tabular} & 196.61  & 17.38\\
\begin{tabular}[c]{@{}c@{}}Trans.+MSF (local)\\+Self-distill (Ours)\end{tabular} & 132.11  & \textbf{10.50}   \\ \hline
\end{tabular}

  Note: 1. Transformer + MSF (MS-output) and Transformer + MSF (local) + self-distill (Ours) denote the combination of Transformer and the explicit outputs of multi-scale fusion blocks and the combination of implicit outputs of multi-scale fusion blocks (\emph{a.k.a.} our MISSU), respectively;
  2. MSF (local) denotes the usage of multi-scale fusion block but with local features for skip connection, while MSF (MS-output) directly uses its output for skip connection;
 3. The input size is $4\times128\times128\times128$ to evaluate GFLOPs and parameters by 5 runnings.

\end{table}

\begin{table*}[t]
 \centering
 \caption{Ablation study for Transformer, multi-scale fusion block (MSF) and self-distillation. Base model is 3D-CNN UNet with all \xmark. }
\label{ab}
\scalebox{0.85}{
\begin{tabular}{cccllclll|cccccc|cc}
\hline
\multirow{3}{*}{Transformer} & \multirow{3}{*}{MSF (local)} & \multicolumn{3}{c}{\multirow{3}{*}{MSF (MS-output)}} & \multicolumn{4}{c|}{\multirow{3}{*}{Self-distill}} & \multicolumn{6}{c|}{BraTS}                                & \multicolumn{2}{c}{CHAOS}                    \\ \cline{10-17} &  & \multicolumn{3}{c}{}& \multicolumn{4}{c|}{} & \multicolumn{3}{c}{$Dice (\%) \uparrow$} & \multicolumn{3}{c|}{$Hausdorff(mm)  \downarrow$} & \multirow{2}{*}{$Dice (\%) \uparrow$} & \multirow{2}{*}{$ACC (\%) \uparrow$} \\ \cline{10-15}
\cline{10-15}& & \multicolumn{3}{c}{}& \multicolumn{4}{c|}{}& $WT$  & $TC$     &$ET$     & $WT$       & $TC$       &$ET$ & & \\ \hline
\xmark & \xmark  & \multicolumn{3}{c}{\xmark} & \multicolumn{4}{c|}{\xmark} & 84.00(1.2)  & 74.00(2.5)  & 71.00(1.7) & 2.4624(2.5)        & 2.0845(3.1)        & 3.2057(2.9)        & 81.14(1.8)                & 81.45(2.8)                   \\
$\pmb{\checkmark}$& \xmark & \multicolumn{3}{c}{\xmark}& \multicolumn{4}{c|}{\xmark}& 86.97(2.8)  & 82.16(3.2)  & 77.45(3.5)  & 1.6653(2.8) &  1.8413(2.0)   &  2.9979(2.1)  & 92.13(3.2) & 93.10(2.9) \\
\xmark& $\pmb{\checkmark}$ & \multicolumn{3}{c}{\xmark} & \multicolumn{4}{c|}{\xmark}& 86.54(2.4)  & 82.21(2.2)  & 77.56(1.4)  & 1.5583(1.2)        & 1.7869(1.7)       & 2.9587(2.6)    & 93.00(1.9)  & 93.86(1.1)\\
\xmark& $\pmb{\checkmark}$ & \multicolumn{3}{c}{\xmark}& \multicolumn{4}{c|}{$\pmb{\checkmark}$} & 87.71(2.1)  & 83.35(1.6)  & 78.82(2.5)  & 1.3870(2.5)  & 1.6984(1.4)        & 2.7892(3.1)& 94.54(2.1) & 94.88(1.8) \\
$\pmb{\checkmark}$ & $\pmb{\checkmark}$  & \multicolumn{3}{c}{\xmark}  & \multicolumn{4}{c|}{\xmark}& 88.63(2.2)  & 84.01(2.4)  & 79.24(1.0)  & 1.0901(2.4) & 1.3400(2.1)& 2.5900(2.5) & 95.28(1.2)& 95.32(1.4)\\
$\pmb{\checkmark}$& \xmark  & \multicolumn{3}{c}{$\pmb{\checkmark}$}& \multicolumn{4}{c|}{\xmark} & \textbf{90.69}(2.2)  & \textbf{86.20}(3.1)  & \textbf{80.73}(2.2)  & \textbf{0.4612}(2.4)& \textbf{0.8099}(2.6)  & \textbf{1.9554}(1.5)  & \textbf{97.09}(3.5)  & \textbf{96.12}(3.1) \\
$\pmb{\checkmark}$  & $\pmb{\checkmark}$ & \multicolumn{3}{c}{\xmark}&\multicolumn{4}{c|}{$\pmb{\checkmark}$}& 89.98(1.3)  & 85.77(2.7)  & 80.14(2.4)  & 0.5008(1.8)        & 0.9423(1.9) & 1.8562(1.8)& 96.78(2.4) & 95.55(2.7) \\ \hline
\end{tabular}}
\end{table*}

\subsection{Ablation Study}

\begin{figure}[t]
 \centering
 \includegraphics[width=1\linewidth]{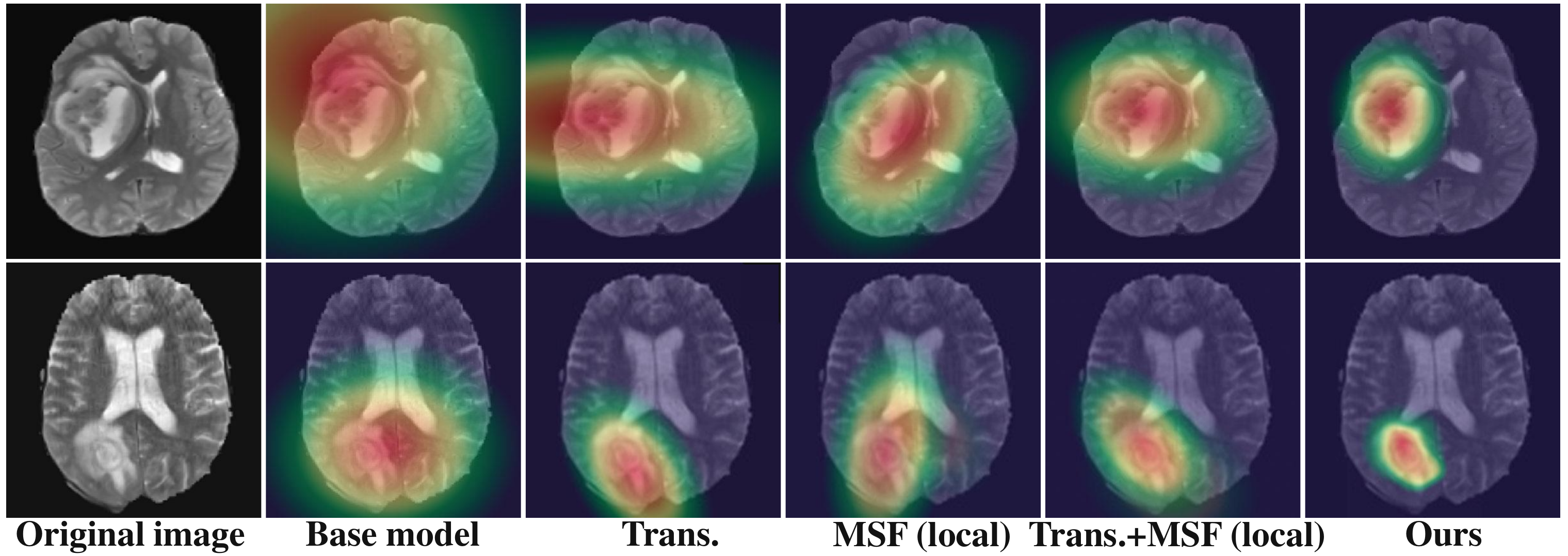}
 \caption{Visual explanations of different components in MISSU for brain tumor segmentation using Grad-CAM. }
 \label{attention}
\end{figure}

\begin{figure}[!t]
 \centering
 \includegraphics[width=1\linewidth]{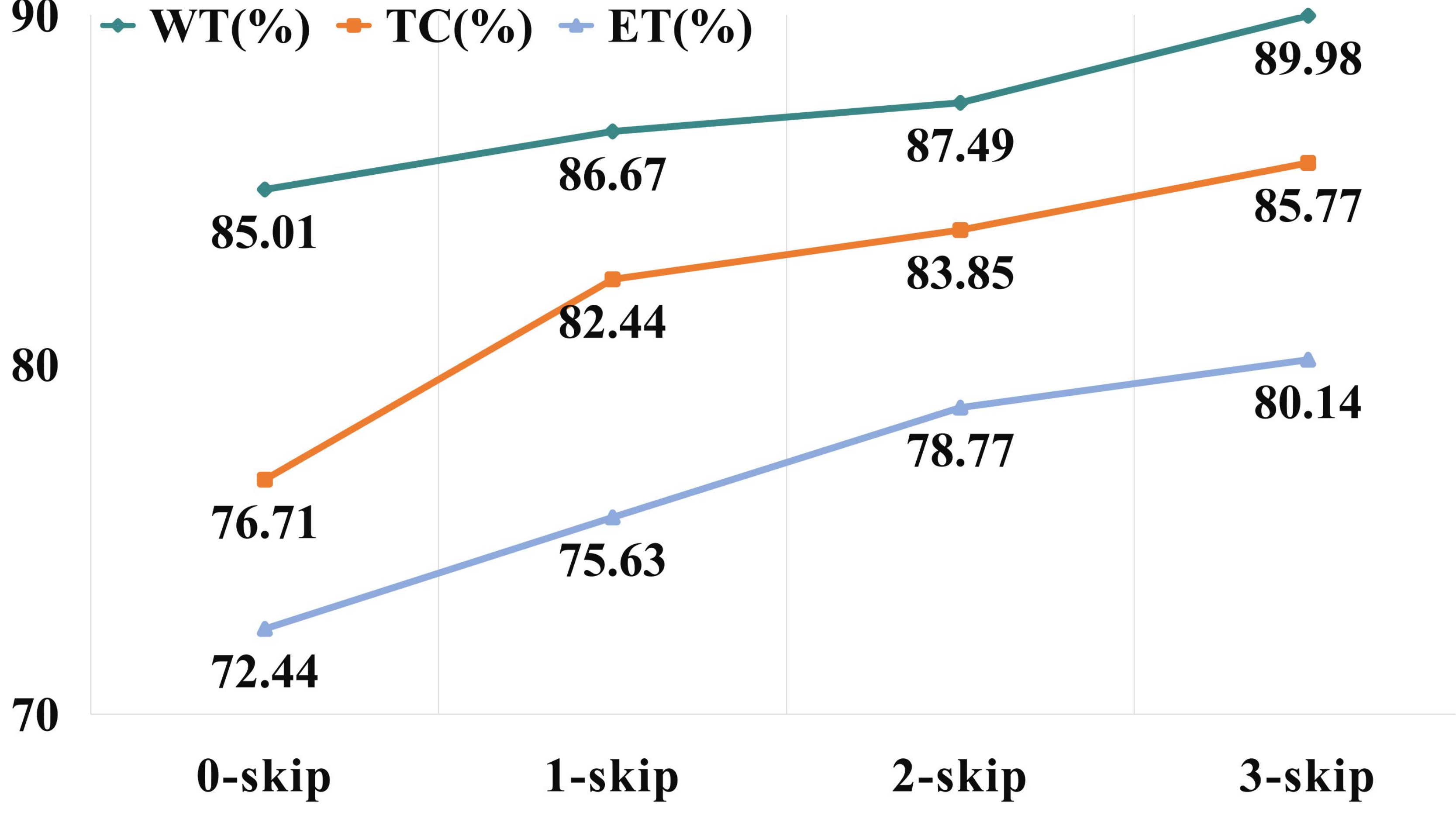}
 \caption{Ablation study of the number of skip-connections on BraTS. }
 \label{skip}
\end{figure}

To verify the effectiveness of MISSU, we conduct ablation studies to analyze different elements, including Transformer, MSF and self-distillation. The results are summarized in Tab.~\ref{ab}, where the base model is \emph{3D-CNN UNet} with all \xmark. Both MSF (local) and MSF (MS-output) use multi-scale fusion blocks, while the former uses local features for skip connection and the latter uses the output by itself for skip connection. 
We also provide qualitative comparison results on the BraTS dataset to effectively demonstrate the effects of different modules, as shown in Fig. \ref{attention}.
Similarly, to validate the performance under different settings of Transformer, a variety of ablation studies were performed, including 1) the number of skip-connections (Fig.~\ref{skip}); 2) Transformer layers $L$ (Tab.~\ref{depdim}) and 3) embedding space $d$ (Tab.~\ref{depdim}).

\textbf{Effect of Transformer.}
We evaluate the effect of the Transformer, (\ie with \textit{vs.} without Transformer). As shown in Tab.~\ref{ab}, compared to the base model, Transformer added after it (\ie the second row) achieves significant performance improvement, especially with the increase of $8.16\%$ and $11.65\%$ ACC on BraTS 2019 and CHAOS dataset, respectively.
Equally, the combination of the Transformer and MSF (local) (\ie the fifth row) achieves a higher Dice-score and Hausdoff compared to only MSF (local) (\ie the third row), the performance of WT gains over $2.97\%$ and $0.80$, respectively. 
We clearly observe that both Transformer and MSF (local) are a necessary in our pipeline. 
It demonstrates the advantages of using Transformer to model global interactions. Furthermore, using the proposed MSF (local) improves MISSU ability to model irregular-shape deformation of lesion regions, while the feature expansion module and the proposed insight of pursuing an inverted bottleneck-like architecture (\ie seeking Transformer width rather than depth) both assist MISSU have richer feature representations.

\textbf{Effect of multi-scale fusion blocks.}
We further evaluate the effectiveness of MSF. Compared to the base model, MSF (local) can learn local detailed features to improve the segmentation performance. For example, adding MSF (local) (\ie the third row) achieves $11.86\%$ Dice score gains over the base model on CHAOS. With Transformer block,
the MSF (local) brings remarkable gain to the Hausdoff scores ($\downarrow 0.58 mm$, $\downarrow 0.50 mm$, $\downarrow 0.41 mm$, separately) and the proposed insight also leads to a great advancement on the Dice score of WT, TC and ET (see the second-row \textit{vs.} the fifth one), jointly enabling MISSU to better solve the inherent problems of medical image segmentation.
Transformer + MSF (MS-output), achieves the best performance, compared to only Transformer + MSF (local). However, it requires the heaviest computation and memory cost for inference, as shown in Tab.~\ref{tab:flop}. Transformer + MSF (local) achieves the reduction of $64.5$ GFLOPs and $6.88$ parameters for four-modality and the number of $128$ slices/images, compared to that of Transformer + MSF (MS-output).
Similarly, we compare the algorithms with 3D UNet and Transformer-based methods (\eg TransUNet, TransBTS and UNETR),
as shown in Tab.~\ref{compare} and Tab.~\ref{tab:flop}, 
MISSU achieves the fastest speed and smallest amount of computation among the compared models, while also achieving very high accuracy.


\textbf{Effect of self-distillation.}
Self-distillation is also an essential element to improve the segmentation performance without extra computation cost at inference.
As can be seen from Table~\ref{ab},
compared to only MSF (local), the self-distillation refines local features to learn more multi-level and detailed features, which achieves at least $1.14\%$ Dice-score improvement on BraTS 2019, \eg $82.21\%$ TC Dice-score in the third row \textit{vs.} $83.35\%$ in the fourth one, and Hausdorff distance also improves of 0.0885mm.
A similar increasing trend by using self-distillation is presented in the fifth and last row. We also find that MSF (MS-output) shown in the sixth row achieves a slightly higher Dice-score than the last row. However, as shown in Tab.~\ref{tab:flop}, the Transformer+MSF (local)+self-distillation (SD) directly reduces $64.5$ GFLOPs and $6.88M$ parameters, which achieves the better trade-off between Dice-score and computation/memory cost.

\textbf{Analysis on visual attention.}  
We further employ gradient-weighted class activation mapping (Grad-CAM)~\cite{selvaraju2017grad} to see the localization attention map from the final convolution layer, which highlights the critical regions of the image. As shown in Fig.~\ref{attention}, the Transformer block tends to learn the global feature presentation, while MSF~(local) focuses on the detailed feature presentation. We leverage MSF into Transformer learned by self-distillation, which significantly concentrates on the brain tumor area for accurate segmentation. It indicates that our method can learn global semantic information and local spatial-detailed features.

\begin{table*}[t]
\centering
\caption{Ablation study on Transformer scale (\eg the number of Transformer layers (\textit{L}) and feature embedding space (\textit{d}).} 
\label{depdim}
\begin{tabular}{cc|cccccc|cc}
\hline
\multirow{3}{*}{Layers ($L$)} & \multirow{3}{*}{\begin{tabular}[c]{@{}c@{}}Embedding \\ space ($d$)\end{tabular}} & \multicolumn{6}{c|}{BraTS}   & \multicolumn{2}{c}{CHAOS}\\ \cline{3-10}  & & \multicolumn{3}{c}{$Dice(\%) \uparrow$} & \multicolumn{3}{c|}{$Hausdorff(mm) \downarrow$} & \multirow{2}{*}{$Dice(\%) \uparrow$} & \multirow{2}{*}{$ACC(\%) \uparrow$} \\ \cline{3-8}& & $WT$       & $TC$      & $ET$      & $WT$      & $TC $      & $ET$       &\\ \hline
1 & 512 & 79.02(3.1)    & 75.34(2.1)   & 71.50(2.7)   & 1.8427(2.0)& 2.2478(1.8)& 2.9133(2.2) &78.99(3.9)&79.11(3.6)    \\
8 & 512 & 80.55(2.9)    & 76.89(1.5)   & 73.58(1.8)   & 1.7207(1.7)        & 2.0986(2.5)      & 2.3585(3.1)&82.33(2.4)&82.96(2.5)       \\
4 & 384 & 83.74(0.9)    & 79.89(1.2)   & 75.01(1.5)   & 1.4501(1.7)        & 1.6233(2.0)      & 2.1783(2.4) &90.06(3.1)&   89.66(3.5)  \\
4 & 512  & \textbf{89.98}(1.3)  & \textbf{85.77}(2.7)    &\textbf{80.14}(2.4)  & \textbf{0.5008}(1.8) & \textbf{0.9423}(1.9) & \textbf{1.8562}(1.8) &\textbf{96.78}(2.4)&\textbf{95.55}(2.7)  \\
4 & 768& 85.15(3.4)    & 80.27(2.6)   & 75.83(2.3)   &1.1555(2.7)         & 1.4993(1.7)       & 2.0966(2.6)&92.18(3.7)&93.19(2.2)       \\ \hline
\end{tabular}
\end{table*}

\textbf{Effect of Transformer scale.}
The scale of Transformer is primarily determined by two hyper-parameters: the feature embedding space ($d$) and the number of Transformer layers ($L$). To verify the impact of Transformer scale on segmentation performance, we perform an ablation study. 
For this purpose, we set the range of the numbers of layers embedding sapce as $\{1, 4, 8\}$ and $\{384, 512, 768\}$, respectively.
As shown in Tab. \ref{depdim}, the performance of our method achieves the best Dice scores of $89.98\%$, $85.77\%$, $80.14\%$ and Hausdorff distance of 0.5008mm, 0.9423mm, 1.8562mm on WT, TC and ET respectively, which are
comparable or higher results than previous SOTA methods.
We can see that MISSU has poor performance $L$ are 1 and 8. 
Increasing the embedding dimension ($d$) may not always result in improved performance ($L = 4$, $d$: 512 vs. 768) yet brings extra computational cost.
The reason can be that the large values of $d$ and $L$ mean the model tends to produce more irrelevant embeddings rather than discriminative embeddings for our tasks,
expanding the embedding space and Transformer layers in a well-designed approach can assist the network reaches higher representation capabilities.

\textbf{Effect of the number of skip-connections}
As previously noted, integrating UNet like skip-connections assists in the enhancement of finer segmentation details by recovering low-level spatial information. The purpose of this ablation is to test the impact of adding different numbers of skip-connections.
By varying the number of skip-connections to be 0/1/2/3, the segmentation performance in Dice on BraTS dataset are summarized in Fig.~\ref{skip}.
We can see that increasing the number of skip-connections improves segmentation performance. Inserting skip-connections to all three intermediate upsampling steps yields the best results.
Due to the recovery of low-level spatial detail information, significant performances were achieved for the critical WT, ET and TC ($89.98\%$, $85.77\%$, $80.14\%$).
These results supported our initial hypothesis that UNet-like skip-connections should be incorporated into the Transformer design to enable learning of precise low-level information.

\section{Discussion}
\label{discussion}

\subsection{Differences from other Transformer-based methods}
To make a thorough comparison between our MISSU and other Transformer-based methods, 
we further give complete analysis to demonstrate the powerful potential of our method.
Although SOTA methods (\emph{e.g.,} UNETR, MeTrans TransUNet and TransBTS) can achieve good performance via CNNs and Transformer, our MISSU has a completely different learning mechanism. 
For example, different from UNETR~\cite{hatamizadeh2021unetr} on directly utilizing the pure Transformer as a backbone to model global and local features. 
We model local-to-global feature interaction in an inverse manner, \emph{i.e.}, first CNN and then transformer. 
Additionally, MeTrans~\cite{jun2021medical} directly models 3D volumetric images in the form of a sequence of 2D image slices, fusing the local features from different layers as input for transformer block, while we progressively refine the local features by the proposed MSF block and use the final local features as input to the transformer layers. 
Similarly, 
both TransUNet~\cite{chen2021transunet} and TransBTS~\cite{wang2021transbts} directly reuse the coarse local features from CNNs to the decoder by the skip connection, while our MISSU transfers the delicate local features using the MSF block and self-distillation into the decoder, which can obtain the multi-scale local detailed features to improve the performance.
Therefore, the proposed local-to-global interaction learning framework in MISSU is novel, especially on 3D local feature enhancement and transfer for the computation-free inference. 
Note that we can employ efficient Transformer variants in our MISSU to reduce memory and compute complexity while preserving accuracy.

 \subsection{The influence of MSF}
 \label{discussion_msf}
 
We make a comparison between MSF (local) and MSF (MS-output), aiming to evaluate the effects of implicit and explicit MSF. 
MSF (local) denotes the multi-scale fusion block adding after the original 3D features but only using the original 3D features for skip connection to the decoder during training, which means the original 3D features are implicitly refined via MSF. Note that MSF (local) does not compute the MSF module for inference.
In contrast, MSF (MS-output) directly uses the outputs of MSF for skip connections during both training and testing, which explicitly uses the refined features and needs more computation for inference. For the comparison, the proposed self-distillation with MSF (local) can capture shape-aware local details, achieving the best trade-off between Dice-score and computation/memory cost, compared to only MSF (local) and MSF (MS-output).
We speculate that the local features generated by MSF may improve the long-range feature modeling of the Transformer, resulting in better performance.

\section{Conclusion}
\label{conclusion}

In this paper, we propose a novel 3D medical image segmentation framework by self-distilling transformer-based UNet (termed \emph{MISSU}), which simultaneously learns global semantic information and local spatial-detailed features. 
In particular, the Transformer block is used to model global semantic context information by adding after the local feature maps from the last 3D CNN layer. Meanwhile, a local multi-scale fusion block is proposed to refine local features extracted by the main CNN stem through self-distillation, which is only computed during training and removed at inference. The global features embedded by Transformer are upsampled and fused into the refined local features in the decoder to generate the pixel-level prediction progressively.
In particular, we analyze the effectiveness of the Transformer, MSF and self-distillation.
We have comprehensively evaluated the performance of MISSU on BraTS 2019 and CHAOS datasets, which demonstrates the superior performance gains over the state-of-the-art methods.
 In the future, we will integrate both MRI and non-image (\eg age and sex) information into our MISSU framework for tumor identification.

 \clearpage


\bibliographystyle{IEEEtran}

\begin{thebibliography}{10}
\providecommand{\url}[1]{#1}
\csname url@samestyle\endcsname
\providecommand{\newblock}{\relax}
\providecommand{\bibinfo}[2]{#2}
\providecommand{\BIBentrySTDinterwordspacing}{\spaceskip=0pt\relax}
\providecommand{\BIBentryALTinterwordstretchfactor}{4}
\providecommand{\BIBentryALTinterwordspacing}{\spaceskip=\fontdimen2\font plus
\BIBentryALTinterwordstretchfactor\fontdimen3\font minus
  \fontdimen4\font\relax}
\providecommand{\BIBforeignlanguage}[2]{{%
\expandafter\ifx\csname l@#1\endcsname\relax
\typeout{** WARNING: IEEEtran.bst: No hyphenation pattern has been}%
\typeout{** loaded for the language `#1'. Using the pattern for}%
\typeout{** the default language instead.}%
\else
\language=\csname l@#1\endcsname
\fi
#2}}
\providecommand{\BIBdecl}{\relax}
\BIBdecl

\bibitem{stoitsis2006computer}
J.~Stoitsis, I.~Valavanis, and S.~G. Mougiakakou, ``Computer aided diagnosis
  based on medical image processing and artificial intelligence methods,''
  \emph{Nuclear Instruments and Methods in Physics Research Section A}, vol.
  569, no.~2, pp. 591--595, 2006.

\bibitem{ronneberger2015u}
O.~Ronneberger, P.~Fischer, and T.~Brox, ``U-net: Convolutional networks for
  biomedical image segmentation,'' in \emph{MICCAI}, 2015, pp. 234--241.

\bibitem{milletari2016v}
F.~Milletari, N.~Navab, and S.-A. Ahmadi, ``V-net: Fully convolutional neural
  networks for volumetric medical image segmentation,'' in \emph{3DV}, 2016,
  pp. 565--571.

\bibitem{wang2018deep}
Y.~Wang, Z.~Deng, X.~Hu, L.~Zhu, X.~Yang, X.~Xu, P.-A. Heng, and D.~Ni, ``Deep
  attentional features for prostate segmentation in ultrasound,'' in
  \emph{MICCAI}, 2018, pp. 523--530.

\bibitem{isensee2021nnu}
F.~Isensee, P.~F. Jaeger, S.~A. Kohl, and Petersen, ``nnu-net: a
  self-configuring method for deep learning-based biomedical image
  segmentation,'' \emph{Nature methods}, vol.~18, no.~2, pp. 203--211, 2021.

\bibitem{wang2021transbts}
W.~Wang, C.~Chen, M.~Ding, H.~Yu, S.~Zha, and J.~Li, ``Transbts: Multimodal
  brain tumor segmentation using transformer,'' in \emph{MICCAI}, 2021, pp.
  109--119.

\bibitem{dosovitskiy2020image}
A.~Dosovitskiy, L.~Beyer, A.~Kolesnikov, D.~Weissenborn, X.~Zhai,
  T.~Unterthiner, M.~Dehghani, Minderer, and Heigold, ``An image is worth
  16$\times$ 16 words: Transformers for image recognition at scale,'' 2020.

\bibitem{chang2021transclaw}
Y.~Chang, H.~Menghan, Z.~Guangtao, and Z.~Xiao-Ping, ``Transclaw u-net: Claw
  u-net with transformers for medical image segmentation,'' \emph{arXiv
  preprint arXiv:2107.05188}, 2021.

\bibitem{mehri2021mprnet}
A.~Mehri, P.~B. Ardakani, and A.~D. Sappa, ``Mprnet: Multi-path residual
  network for lightweight image super resolution,'' in \emph{WCACV}, 2021, pp.
  2704--2713.

\bibitem{dolz2018hyperdense}
J.~Dolz, K.~Gopinath, J.~Yuan, H.~Lombaert, C.~Desrosiers, and I.~B. Ayed,
  ``Hyperdense-net: a hyper-densely connected cnn for multi-modal image
  segmentation,'' \emph{TMI}, vol.~38, no.~5, pp. 1116--1126, 2018.

\bibitem{chen2018drinet}
L.~Chen, P.~Bentley, K.~Mori, K.~Misawa, M.~Fujiwara, and D.~Rueckert, ``Drinet
  for medical image segmentation,'' \emph{TMI}, vol.~37, no.~11, pp.
  2453--2462, 2018.

\bibitem{farabet2012learning}
C.~Farabet, C.~Couprie, L.~Najman, and Y.~LeCun, ``Learning hierarchical
  features for scene labeling,'' \emph{TPAMI}, vol.~35, no.~8, pp. 1915--1929,
  2012.

\bibitem{feng2020cpfnet}
S.~Feng, H.~Zhao, F.~Shi, X.~Cheng, M.~Wang, Y.~Ma, D.~Xiang, W.~Zhu, and
  X.~Chen, ``Cpfnet: Context pyramid fusion network for medical image
  segmentation,'' \emph{TMI}, vol.~39, no.~10, pp. 3008--3018, 2020.

\bibitem{kamnitsas2017efficient}
K.~Kamnitsas, C.~Ledig, V.~F. Newcombe, J.~P. Simpson, A.~D. Kane, D.~K. Menon,
  D.~Rueckert, and B.~Glocker, ``Efficient multi-scale 3d cnn with fully
  connected crf for accurate brain lesion segmentation,'' \emph{MIA}, vol.~36,
  pp. 61--78, 2017.

\bibitem{han2020attention}
Y.~Han, J.~Li \emph{et~al.}, ``An attention-oriented u-net model and global
  feature for medical image segmentation,'' \emph{Journal of Applied Science
  and Engineering}, vol.~23, no.~4, pp. 731--738, 2020.

\bibitem{schlemper2019attention}
J.~Schlemper, O.~Oktay, M.~Schaap, M.~Heinrich, B.~Kainz, B.~Glocker, and
  D.~Rueckert, ``Attention gated networks: Learning to leverage salient regions
  in medical images,'' \emph{MIA}, vol.~53, pp. 197--207, 2019.

\bibitem{zhou2019models}
Z.~Zhou, V.~Sodha, M.~M.~R. Siddiquee, R.~Feng, N.~Tajbakhsh, M.~B. Gotway, and
  J.~Liang, ``Models genesis: Generic autodidactic models for 3d medical image
  analysis,'' in \emph{MICCAI}, 2019, pp. 384--393.

\bibitem{wang2018non}
X.~Wang, R.~Girshick, A.~Gupta, and K.~He, ``Non-local neural networks,'' in
  \emph{CVPR}, 2018, pp. 7794--7803.

\bibitem{taleb2021multimodal}
A.~Taleb, C.~Lippert, T.~Klein, and M.~Nabi, ``Multimodal self-supervised
  learning for medical image analysis,'' in \emph{International Conference on
  Information Processing in Medical Imaging}, 2021, pp. 661--673.

\bibitem{tao2020revisiting}
X.~Tao, Y.~Li, W.~Zhou, K.~Ma, and Y.~Zheng, ``Revisiting rubik’s cube:
  Self-supervised learning with volume-wise transformation for 3d medical image
  segmentation,'' in \emph{MICCAI}, 2020, pp. 238--248.

\bibitem{li2021rotation}
X.~Li, X.~Hu, X.~Qi, L.~Yu, W.~Zhao, P.-A. Heng, and L.~Xing,
  ``Rotation-oriented collaborative self-supervised learning for retinal
  disease diagnosis,'' \emph{TMI}, 2021.

\bibitem{chen2019self}
L.~Chen, P.~Bentley, K.~Mori, K.~Misawa, M.~Fujiwara, and D.~Rueckert,
  ``Self-supervised learning for medical image analysis using image context
  restoration,'' \emph{MIA}, vol.~58, p. 101539, 2019.

\bibitem{ross2018exploiting}
T.~Ross, D.~Zimmerer, A.~Vemuri, F.~Isensee, and Wiesenfarth, ``Exploiting the
  potential of unlabeled endoscopic video data with self-supervised learning,''
  \emph{International journal of computer assisted radiology and surgery},
  vol.~13, no.~6, pp. 925--933, 2018.

\bibitem{korkmaz2022unsupervised}
Y.~Korkmaz, S.~U. Dar, M.~Yurt, M.~{\"O}zbey, and T.~Cukur, ``Unsupervised mri
  reconstruction via zero-shot learned adversarial transformers,'' \emph{TMI},
  2022.

\bibitem{hatamizadeh2021unetr}
A.~Hatamizadeh, D.~Yang, Roth, and D.~Xu, ``Unetr: Transformers for 3d medical
  image segmentation,'' \emph{arXiv preprint arXiv:2103.10504}, 2021.

\bibitem{chen2021transunet}
J.~Chen, Y.~Lu, Q.~Yu, X.~Luo, E.~Adeli, Y.~Wang, L.~Lu, A.~L. Yuille, and
  Y.~Zhou, ``Transunet: Transformers make strong encoders for medical image
  segmentation,'' \emph{arXiv preprint arXiv:2102.04306}, 2021.

\bibitem{chandrakar2020brain}
M.~K. Chandrakar and A.~Mishra, ``Brain tumor detection using multipath
  convolution neural network,'' \emph{IJCVIP}, vol.~10, no.~4, pp. 43--53,
  2020.

\bibitem{wei2020multi}
J.~Wei, G.~Lin, K.-H. Yap, T.-Y. Hung, and L.~Xie, ``Multi-path region mining
  for weakly supervised 3d semantic segmentation on point clouds,'' in
  \emph{CVPR}, 2020, pp. 4384--4393.

\bibitem{wu2019vessel}
Y.~Wu, Y.~Xia, Y.~Song, D.~Zhang, D.~Liu, C.~Zhang, and W.~Cai, ``Vessel-net:
  retinal vessel segmentation under multi-path supervision,'' in \emph{MICCAI},
  2019, pp. 264--272.

\bibitem{zhao2019automated}
J.~Zhao, Q.~Li, X.~Li, H.~Li, and L.~Zhang, ``Automated segmentation of
  cervical nuclei in pap smear images using deformable multi-path ensemble
  model,'' in \emph{ISBI}, 2019, pp. 1514--1518.

\bibitem{qin2021efficient}
D.~Qin, J.-J. Bu, Z.~Liu, X.~Shen, S.~Zhou, J.-J. Gu, Z.-H. Wang, L.~Wu, and
  H.-F. Dai, ``Efficient medical image segmentation based on knowledge
  distillation,'' \emph{TMI}, vol.~40, no.~12, pp. 3820--3831, 2021.

\bibitem{zhang2021cross}
G.~Zhang, X.~Shen, Y.~Zhang, Y.~Luo, J.~Luo, D.~Zhu, H.~Yang, W.~Wang, B.~Zhao,
  and J.~Lu, ``Cross-modal prostate cancer segmentation via self-attention
  distillation,'' \emph{IEEE Journal of Biomedical and Health Informatics},
  2021.

\bibitem{ahn2019variational}
S.~Ahn, S.~X. Hu, A.~Damianou, N.~D. Lawrence, and Z.~Dai, ``Variational
  information distillation for knowledge transfer,'' in \emph{CVPR}, 2019.

\bibitem{yang2019training}
C.~Yang, L.~Xie, S.~Qiao, and A.~L. Yuille, ``Training deep neural networks in
  generations: A more tolerant teacher educates better students,'' in
  \emph{AAAI}, 2019, pp. 5628--5635.

\bibitem{hou2019learning}
Y.~Hou, Z.~Ma, C.~Liu, and C.~C. Loy, ``Learning lightweight lane detection
  cnns by self attention distillation,'' in \emph{ICCV}, 2019.

\bibitem{kang2020kd}
J.~Kang and J.~Gwak, ``Kd-resunet++: Automatic polyp segmentation via
  self-knowledge distillation,'' 2020.

\bibitem{yuan2021tokens}
L.~Yuan, Y.~Chen, T.~Wang, W.~Yu, Y.~Shi, Z.~Jiang, and Tay, ``Tokens-to-token
  vit: Training vision transformers from scratch on imagenet,'' in \emph{CVPR},
  2021.

\bibitem{chen2017rethinking}
L.-C. Chen, G.~Papandreou, F.~Schroff, and H.~Adam, ``Rethinking atrous
  convolution for semantic image segmentation,'' \emph{arXiv preprint
  arXiv:1706.05587}, 2017.

\bibitem{lin2017feature}
T.-Y. Lin, P.~Doll{\'a}r, R.~Girshick, K.~He, B.~Hariharan, and S.~Belongie,
  ``Feature pyramid networks for object detection,'' in \emph{CVPR}, 2017, pp.
  2117--2125.

\bibitem{komodakis2017paying}
N.~Komodakis and S.~Zagoruyko, ``Paying more attention to attention: improving
  the performance of convolutional neural networks via attention transfer,'' in
  \emph{ICLR}, 2017.

\bibitem{kingmaadam}
D.~P. Kingma and J.~L. Ba, ``Adam: Amethod for stochastic optimization.''

\bibitem{bakas2018identifying}
S.~Bakas, M.~Reyes, E.~Battistella, S.~Chandra, T.~Estienne, L.~Fidon, and
  Vakalopoulou, ``Identifying the best machine learning algorithms for brain
  tumor segmentation, progression assessment, and overall survival prediction
  in the brats challenge,'' 2018.

\bibitem{kavur2019chaos}
A.~E. Kavur, M.~A. Selver, O.~Dicle, M.~Bar{\i}s, and N.~S. Gezer,
  ``Chaos-combined (ct-mr) healthy abdominal organ segmentation challenge
  data,'' in \emph{ISBI}, 2019.

\bibitem{frangi2018medical}
A.~F. Frangi, J.~A. Schnabel, C.~Davatzikos, and Alberola-L{\'o}pez, ``Medical
  image computing and computer assisted intervention--miccai 2018,'' in
  \emph{MICCAI}, vol. 11073, 2018, p. 534.

\bibitem{cciccek20163d}
{\"O}.~{\c{C}}i{\c{c}}ek, A.~Abdulkadir, S.~S. Lienkamp, T.~Brox, and
  O.~Ronneberger, ``3d u-net: learning dense volumetric segmentation from
  sparse annotation,'' in \emph{MICCAI}, 2016, pp. 424--432.

\bibitem{hu2020knowledge}
M.~Hu, M.~Maillard, Y.~Zhang, T.~Ciceri, and L.~Barbera, ``Knowledge
  distillation from multi-modal to mono-modal segmentation networks,'' in
  \emph{MICCAI}, 2020, pp. 772--781.

\bibitem{kayal2020region}
S.~Kayal and S.~Chen, ``Region-of-interest guided supervoxel inpainting for
  self-supervision,'' in \emph{MICCAI}, 2020, pp. 500--509.

\bibitem{jun2021medical}
E.~Jun, S.~Jeong, D.-W. Heo, and H.-I. Suk, ``Medical transformer: Universal
  brain encoder for 3d mri analysis,'' \emph{arXiv preprint arXiv:2104.13633},
  2021.

\bibitem{sinha2020multi}
A.~Sinha and J.~Dolz, ``Multi-scale self-guided attention for medical image
  segmentation,'' \emph{IEEE journal of biomedical and health informatics},
  2020.

\bibitem{selvaraju2017grad}
R.~R. Selvaraju, M.~Cogswell, A.~Das, R.~Vedantam, D.~Parikh, and D.~Batra,
  ``Grad-cam: Visual explanations from deep networks via gradient-based
  localization,'' in \emph{ICCV}, 2017, pp. 618--626.

\end{thebibliography}


\end{document}